%% file: main.tex
\documentclass{article}

\usepackage{times}
\usepackage{soul}
\usepackage{url}
\usepackage[utf8]{inputenc}
\usepackage[small]{caption}
\usepackage{graphicx}
\usepackage{amsmath}
\usepackage{amsthm}
\usepackage{booktabs}
\usepackage{algorithm}
\usepackage{algorithmic}
\usepackage{graphicx}
\usepackage{stmaryrd} 
\usepackage{amssymb}
\usepackage{multirow}
\usepackage[table]{xcolor}
\usepackage{color}
\usepackage{comment}
\usepackage{cite}
\usepackage{microtype}
\usepackage{graphicx}
\usepackage{subfigure}
\usepackage{booktabs} 
\usepackage{tikz}
\usepackage{smartdiagram}
\usepackage{comment}

\newcommand{\fakepar}[1]{~\\\noindent\textit{#1.}}
\renewcommand{\fakepar}[1]{\subsection{#1}}
\newcommand{\censure}[1]{}

\usepackage{hyperref}
\usepackage[accepted]{icml2021}

\icmltitlerunning{Thinkback: Task Specific Out-of-Distribution Detection.}

\begin{document}

\twocolumn[
\icmltitle{\huge{Thinkback:} Task-Specific\\ Out-of-Distribution Detection}




\begin{icmlauthorlist}
\icmlauthor{Lixuan YANG}{to}
\icmlauthor{Dario ROSSI}{to}
\end{icmlauthorlist}

\icmlaffiliation{to}{Huawei Technologies, France}

\icmlcorrespondingauthor{Lixuan YANG}{lixuan.yang@huawei.com}
\icmlcorrespondingauthor{Dario ROSSI}{dario.rossi@huawei.com}

\icmlkeywords{Openset Recognition, Out-of-distribution detection, Task specific detection}

\vskip 0.3in
]



\printAffiliationsAndNotice{} 

\begin{abstract}
The increased success of  Deep Learning (DL) has recently sparked large-scale deployment of DL models in many diverse industry segments.
Yet, a crucial weakness of  supervised model is the inherent difficulty in handling  out-of-distribution samples, i.e., samples belonging to classes that were not presented to the model at training time.

We propose in this paper a novel way to formulate the out-of-distribution detection problem, tailored for DL models. Our method does not require  fine tuning process on training data,  yet is significantly  more  accurate than the state of the art for out-of-distribution detection. 
\end{abstract}

\section{Introduction}
\label{sec:intro}
Irrespectively of the specific application, classes and inputs, classification engines needs to perform two functions: namely an \emph{identification function} $f(x)$  and an \emph{ open-set detection function} $g(x)$. Shortly, the goal of $\ell = f(x)$ is to  determine from an input $x$, a label $\ell\in[1,K]$ among a set of $K$ known classes. 
Supervised ML/DL techniques are well suited to learn the function $f(x)$, in a process referred to as training, where 
the parameters $W$ of the function $f(x)$ are adapted (e.g., through backpropagation in case of DL models).  The goal of $g(x)$ is instead to detect  open-set inputs $x'$, i.e., inputs that do not belong to any of the $K$ classes known to $f(x)$.

The problem tackled by $g(x)$ is usually referred to as \emph{out of distribution} (OOD) detection, and has recently gathered significant attention in the literature for a variety of domains, ranging from network\cite{Yang2021}, natural language processing\cite{Miok2020} to medical field~\cite{Cao2020}.
Given this variety of applications domains, ideally OOD techniques should be broadly applicable to any DL model deployed in production, and require as few assistance as possible.
Additionally, OOD techniques computational complexity should be small, as if  $g(x)$ is slower than $f(x)$, then this would either limit the classification rate of $f(x)$, or means that OOD could be applied only to some samples, i.e., trading off classification rate performance with OOD detection accuracy.  
Conscious of the above constraints, in this paper we propose ``Thinkback'', a novel gradient-based method for task-specific OOD that does not require any specific tuning, is computationally very lightweight, yet very accurate. In a nutshell, after having classified a new instance $x$ with the feed-forward network inference $f(x)=\ell$,  our method infer the plausibility of $\ell$ by pretending $\ell$ to be a ground truth label, and assessing how much this new label would change model weights, by performing the initial steps of a backpropagation -- where we expect weights to change little if $\ell$ is a plausible class for $x$, and change much in case $x$ is an OOD sample.

In the reminder of this short paper, we overview the state of the art (Sec.~\ref{sec:related}), we present  (Sec.~\ref{sec:system}) our Thinkback method
and benchmark (Sec.~\ref{sec:evaluation}) our proposed Thinkback with the state of the art methods and summarize our findings  (Sec.~\ref{sec:discussion}).




\section{Related work}\label{sec:related}

\input{02_related}


\section{Methodology}\label{sec:system}
\input{04_system}

\section{Evaluation}\label{sec:evaluation}
\input{05_evaluation}

\section{Discussion}\label{sec:discussion}
\input{07_discussion}

\clearpage

\bibliography{main}
\bibliographystyle{icml2021}


\end{document}

%% file: 02_related.tex
We summarize the relevant literature with the help of  Fig.~\ref{fig:related}, that depicts a visual synoptic where related work is divided into three categories, based on whether OOD detection is performed on input $x$, output $\ell=f(x)$ or in the inner-stages of the DL model. 

\input{02_related_figure}

\fakepar{Input}
Works have recently proposed to apply transformations on the input so to control models outputs regulated by mechanisms such as Mahalanobis based score~\cite{Lee18} or temperature scaled SoftMax score~\cite{Liang17}. There are two drawback of those proposals: i) their known additional computational cost makes them unappealing from practical perspective and ii) the fact that input transformation should be concerted prior to DL model deployment, for which work in the remaining classes is more interesting.

 \fakepar{Inner}
At their core, DL methods project input data into a \emph{latent space} where is easier to separate data based on class labels. A set of work then proposes specific ways to alter  this  latent space to purposely simplify open-set recognition.

For instance, ~\cite{Zhao19}  uses AutoEncoders (AE) to transform input data, and apply clustering to the transformed input, while \cite{Yoshihashi19}   uses latent representation along with OpenMax~\cite{Bendale15} activation vectors. Other works instead rely on Generative Adversarial Networks (GAN) to explore the latent space in order to generate ``unknown classes'' data to train a classifier for the class $\ell=0$ i.e., a K+1 classifier. For instance,  \cite{Ge17}  generate unknown classes by mixing the latent representation of known classes,
while ~\cite{Neal2018}  uses optimisation methods to create counterfactual samples that are close to training samples but do not belong to training data. All these methods require specific architectures (so they are hardly deployable) and  extra training (so their computational complexity can be high).

Other work propose to alter activation ~\cite{Jang20} or  loss functions ~\cite{Hassen18, Aljalbout18}. 
In~\cite{Jang20} authors replace the SoftMax activation with a sigmoid, and fit a Weibull distribution for each activation output to revise the output activation.
Special clustering loss functions~\cite{Hassen18, Aljalbout18} can be used to further constraint points of the same class to be close to each other, so that unknown classes are expected to be projected into sparse region which is far from known classes.
However, all these methods constrain to use special DL architectures and cannot be used on existing models; additionally such architectural modifications can alter the accuracy of the supervised classification task. As such, we deem it difficult for techniques of this class to actually broadly deployable, and we disregard them in what follows. 




\fakepar{Output}
The most common approach for OOD detection at output stage is thresholding SoftMax values~\cite{Hendrycks2017}. OpenMax~\cite{Bendale15}  
revises SoftMax activation vectors adding a special ``synthetic'' unknown class, by using weighting induced by Weibull modeling of input data. Alternative approaches include the use of Extreme Value Machine (EVM)~\cite{Rudd15},  based of Extreme Value Theory (EVT), 
and, more recently, clustering on the CNN feature vectors with a PCA reduction of dimension~\cite{Zhang20}.  The authors~\cite{Wang2020} adjusts to few shot classification by using SPP~\cite{Hendrycks2017} and Mahalanobis distance \cite{Lee18}.
 Finally, as a collection of energy values could be turned into a probability density though Gibbs distribution, ~\cite{liu2021energybased} formulates the energy function by the denominator of the SoftMax activation, so that energy scores align with the probability density. This method shows nearly perfect performance after fine tuning on the OOD data -- however, in reality collecting OOD data in advance is impossible.  

Our proposed methods, based on evaluating gradients change via backpropagation,
also  fit in this class: this makes techniques in this class  particularly relevant and worth considering for a direct performance comparison.

We select the best-in class approach, that represents the state of the art and namely Energy score ~\cite{liu2021energybased}, that authors shows to have superior performance to \cite{Liang17,Lee18,Dan2019}.
To better assess the added value of Energy score and our method,  we additionally consider classic  SoftMax outputs~\cite{Hendrycks2017} as a reference.  Based on a preliminary evaluation on additional techniques that were not considered in ~\cite{liu2021energybased},  we
instead discard OpenMax~\cite{Bendale15}, which we find to be significantly less accurate than Energy score. We also discard ~\cite{Zhang20}, which in reason of the PCA and  clustering step, is significantly  more complex from a computational standpoint. Thus, we resort to comparing Energy score, SoftMax and our proposed Thinkback method, for which accuracy comparison can be done on a fair ground in terms of complexity.

%% file: 02_related_figure.tex
\usetikzlibrary{arrows,shadows,positioning}

\tikzset{
  frame/.style={
    rectangle, draw, 
    text width=6em, text centered,
    minimum height=4em,drop shadow,fill=yellow!40,
    rounded corners,
  },
  line/.style={
    draw, -latex',rounded corners=3mm,
  }
}

\begin{figure}
\centering
\resizebox{\columnwidth}{!}{%

\begin{tikzpicture}[font=\small\sffamily\bfseries,very thick,node distance = 2cm]

\node[inner sep=0pt] (cnn) at (0,0)
    {\includegraphics[width=\textwidth]{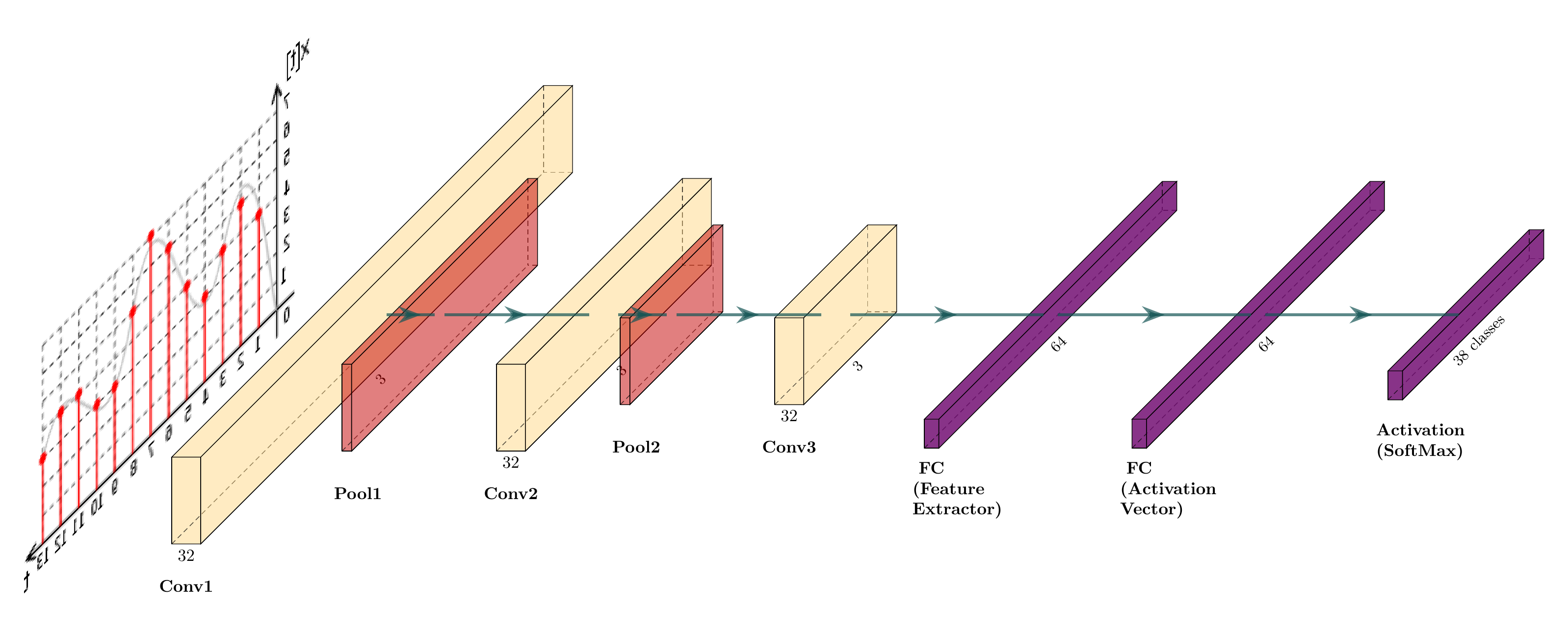}};
\node[frame] (C) at (-5,4) [draw,thick,minimum width=1cm,minimum height=1cm] {Input Preprocessing \cite{Zhang15,Liang17,Lee18}};

\node[frame] (D) at (2,4) [draw,thick,minimum width=1cm,minimum height=1cm] {Loss\\Function\cite{Hassen18}};
\node[frame] (F) at (4,-5) [draw,thick,minimum width=1cm,minimum height=1cm] {Classifier \\ K+1 \cite{Neal2018}};
\node[frame] (K) at (5,4) [draw,thick,minimum width=1cm,minimum height=1cm] {Softmax \cite{Hendrycks2017, Liang17}};

\node[frame] (B) at (-3,-3.8) [draw,thick,minimum width=1cm,minimum height=1cm] {Autoencoder};
\node[frame] (A) at (0,-5) [draw,thick,minimum width=1cm,minimum height=1cm] {GAN};

\node[frame] (L) at (0,-3) [draw,thick,minimum width=1cm,minimum height=1cm] {Reconstruction Error}; 

\node[frame] (I) at (6.5,-3) [draw,thick,minimum width=0.2cm,minimum height=1cm] {OpenMax\\ \cite{Bendale15}};

\node[frame] (J) at (6.5,-5) [draw,thick,minimum width=2cm,minimum height=1cm] {Mahalanobis\\\cite{Lee18} \\ Clustering\\\cite{Zhang20}};
\node[frame] (G) at (8,4) [draw,thick,minimum width=1cm,minimum height=1cm] {Gradient\\based\cite{Yang2021}};

\path [line] (C) -- node[right,align=left,pos=.5] {Augment\\Input}(-5, 1);
\path [line] (-5, -2) |- (B) ;
\path [line] (B) |- (A) ;
\path [line] (D.350) -- (7, 0) ;
\path [line] (G) -- (7, 0) ;
\path [line] (K) -- (7, 0) ;
\path [line] (B) |- node[right,align=right,pos=.7] {Latent \\[3mm] Repr.} (L) ;
\path [line] (A) -- node[right,align=left,pos=.1] {Generate \\[3mm]Unknowns} (F) ;
\path [line] (J) -- (2, -1) ;
\path [line] (I) -- (4, -1) ;

\end{tikzpicture}
} 
\caption{Synoptic of OOD detection wrt an exemplary convolutional DL model: we classify OOD techniques as either working at the input/output of the DL model, or by requiring  modifications to the inner models.} \label{fig:related}

\end{figure}
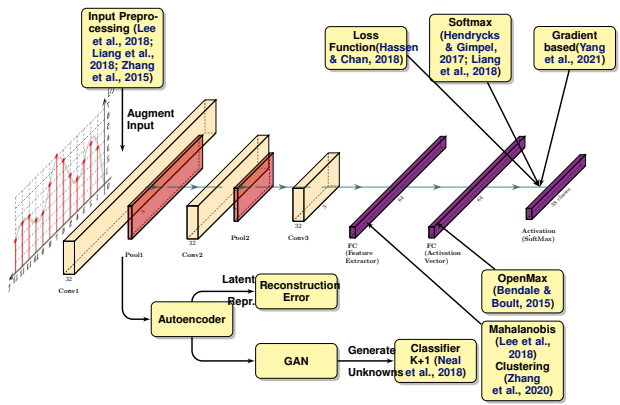

%% file: 04_system.tex
Since the unknown data may take infinite forms, modeling unknown data without any assumption is difficult;  conversely, constraining unknown data by specific assumptions may results in weak OOD detection capabilities in the general case. 

To circumvent this conudrum, we let a DL model  trained on $D_{in}^{train}$ to  \emph{introspect} its feed-forward decision based on the information  extracted  solely out of model weights $W$.  In a nutshell, we ask the model to \emph{think back} about its feed-forward decision by leveraging \emph{backpropagation}, assessing the magnitude of the change that would happen to model weights if they were altered $W'$ by training also on test data $D_{test}$. Thus, we transform the OOD problem into  compute the probability $p(W' | D_{in}^{train})$ of getting $W'$ with a model trained on $D_{in}^{train}$, 
as we detail next.

\subsection{Task specific novelty detection}
\label{sec:sys:task}
In more details, Elastic weight consolidation \cite{Kirkpatrick2016} approximates the posterior probability of $p(w_i|D_{train})$ as a Gaussian distribution with mean given by $\mu_i$ and a diagonal precision $p_i$ given  by  the  diagonal  of the Fisher information matrix $F$. After training on $D_{in}^{train}$, the weights $w_i$ are stable around the $\mu_i$. The probability of a weight, given the training dataset $D_{in}^{train}$ can be written as:
\begin{equation}
p(w_i | D_{in}^{train})=\sqrt{\frac{F_i}{2\pi}} \exp{(-\frac{1}{2}(w_i-\mu_i)^2F_i))}
\end{equation}
When new data $x\in D_{test}$ comes, in addition to performing feed-forward inference $\ell = f(x)$,  we let the model think back about this decision. To do so, we initiate a (fake) backpropagation step, as if the classification outcome $\ell$ was a ground truth label (but without altering the model weights):  the new data would induce changes on model weights as $(w_i)' = w_i - \delta w_i$. We then assess the probability of this newly changed weight when $x$ is an in-distribution sample, which can be written as:
\begin{equation}
p({w_i}' | D_{in}^{train})=\sqrt{\frac{F_i}{2\pi}} \exp{(-\frac{1}{2}(\delta w_i)^2F_i))}
\end{equation}
\noindent Since the weights distribution are independent, we have:
\begin{equation}
p({w_1}', {w_2}', ..., {w_n}'|D_{IN}^{train}) = \prod_i p({w_i}'|D_{in}^{train})
\end{equation}
so that the whole weights distribution is proportional to:
\begin{equation}
p(W'|D_{in}^{train}) \propto \prod_i \exp\left(-\frac{1}{2}(\delta w_i)^2F_i\right) \propto -\sum_i (\delta w_i)^2
\end{equation}

To limit computation complexity, we limit backpropagation to the penultimate layer,  which  contains most information concerning the classes. The intuition is that shall the new sample belong to an OOD class that was never seen at training,  weights of the penultimate  layer which is task-specific should have large changes. Ultimately,  we define the  unknown level of a sample $x$  as the reverse of the above probability:
\begin{equation}
OOD(x \in D_{out}^{test}) \propto \sum_i (\delta w_i)^2
\end{equation}
While our method do not require to fine-tune on OOD data, as for instance \cite{liu2021energybased} (which is however unavailable at training time) it can however benefit from in-distribution training data (which is easily accessible at training time). In particular, during the training phase, despite the local minimum has been reached, the training data still generate gradients. In order to reduce the gradient that follows the in-distribution training data's gradient trend, the gradient is divided by the expected gradient of the training.
\begin{equation}
OOD(x \in D_{out}^{test}) = \sum_i \frac{(\delta w_i)^2}{\epsilon + E(\delta w_i^2|D_{in}^{train})}
\end{equation}
\noindent where $\epsilon$ is a technicality to avoid division by zero in the rescaling. Additionally, as suggested by ODIN~\cite{Liang17}, temperature scaled Softmax score helps to separate the in- and out-of-distribution: we thus backpropagate the scaled Softmax by temperature $T$. 
\begin{equation}
\delta w_i = \frac{\partial L}{\partial w_i} = - \frac{ \partial \sum_{i=1}^{K} y_i \log(Softmax(z_i/T)) }{\partial w_i}
\end{equation}
 By trusting the network's prediction $y_i$, the gradient is the partial derivative of the loss function $L$ on the network softmax activation $z_i$ scaled by $T$ over the total number of classes $K$. 


%% file: 05_evaluation.tex
\newcommand{\fnsize}[1]{\begin{scriptsize}#1\end{scriptsize}}

\begin{table*}[!t]
\begin{center}
\begin{footnotesize}
\centering
    \caption{OOD detection performance of Thinkback, our proposed gradient-based method, compared with vanilla Softmax~\cite{Hendrycks2017} and state of the art Energy Score~\cite{liu2021energybased}. $\uparrow, +$ denote metrics for which larger values are better, while  $\downarrow, -$  indicate that smaller values are better. $\Delta$ shows the relative performance w.r.t Softmax.}\label{tab:accuracy}
    \begin{tabular}{llcccccccc} 
\toprule
\multirow{2}{*}{\bf Dataset} & \multirow{2}{*}{\bf Method}  
    & \bf TPR10 & $\Delta${\bf TPR10} & \bf FPR95 & $\Delta${\bf  FPR95} & \bf AUROC & $\Delta${\bf AUROC} & \bf AUPR & $\Delta$ {\bf AUPR}\\
    && \fnsize{$\uparrow$} & \fnsize{$+$} & \fnsize{$\downarrow$} & \fnsize{$-$} & \fnsize{$\uparrow$} & \fnsize{$+$} & \fnsize{$\uparrow$}  \fnsize{$+$}\\
\midrule    
\multirow{3}{*}{\shortstack{Mean\\Result}}
                    & Softmax   &71.83 & &30.81 & &91.43 & &66.45 &\\
                    & Energy & 80.51 &\emph{+8.68\%} & 36.62 &\emph{+5.81\%} & 92.32&\emph{+0.89\%} & 75.75& \emph{+9.30\%}\\
                    & Thinkback & \bf 84.37&\emph{+12.54\%} & \bf 22.65 &\emph{-8.16\%} & \bf 94.17&\emph{+2.74\%} & \bf 92.42 & \emph{+25.97\%}\\
\midrule
\multirow{3}{*}{Textures}
                    & Softmax& 62.00 & &43.23 && 88.66 && 59.04  &\\
                    & Energy &  65.15 &\emph{+3.15\%} & 68.79&\emph{+25.56\%} & 85.03 &\emph{-3.63\%} & 59.80&\emph{+0.76\%}\\
                    & Thinkback & \bf 78.23 &\emph{+16.23\%} & \bf 32.09 &\emph{-11.14\%} & \bf 92.35 &\emph{+3.69\%} & \bf 91.73&\emph{+32.69\%}\\
\midrule    
\multirow{3}{*}{SVHN}
                    & Softmax  & 71.30    &                & 28.47& & 91.83& & 66.54&  \\
                    & Energy   & 78.25    & \emph{+6.95\%} & 42.29 &\emph{+13.82\%} & 91.07&\emph{-0.76\%} & 70.87&\emph{+4.37\%}\\
                    & Thinkback & \bf 83.82& \emph{+12.52\%} & \bf 23.41& \emph{-5.06\%} & \bf 94.06& \emph{+2.23\%} & \bf 91.90& \emph{+25.4\%}\\
\midrule            
\multirow{3}{*}{\shortstack{LSUN\\Crop}}
                    & Softmax  &  89.15 &              & 15.18 & & 95.64& & 79.78 & \\
                    & Energy   & \bf 96.35    & \emph{+7.20\%}& \bf 6.72 & \emph{-8.46\%} & \bf 98.44  & \emph{+2.90\%} & \bf 93.55 & \emph{+13.77\%}\\
                    & Thinkback & 89.14 & \emph{-0.01\%}& 14.09 & \emph{-1.09\%} & 95.12 & \emph{-0.52\%} & 91.86& \emph{+12.08\%}\\
\midrule            
\multirow{3}{*}{\shortstack{LSUN\\Resize}}
                    & Softmax  & 70.20 & & 30.80& & 91.27& & 64.88 &\\
                    & Energy  & 83.25 &\emph{+13.05\%} & 29.96 &\emph{-0.83\%} & 93.99 &\emph{+2.72\%} & 78.93 & \emph{+14.05\%}\\
                    & Thinkback & \bf 87.86 &\emph{+17.66\%}&\bf 20.73 &\emph{-10.07\%} & \bf 94.88 & \emph{+3.61\%} & \bf 93.47&\emph{+28.59}\%\\
\midrule            
\multirow{3}{*}{iSUN}
                    & Softmax    & 66.50 && 36.37 && 89.76& & 61.99 &\\
                    & Energy  & 79.55 &\emph{+13.05\%} & 35.34&\emph{-1.03\%} & 93.06&\emph{+3.30\%} & 75.63&\emph{+13.64\%}\\
                    & Thinkback & \bf 82.82 & \emph{+16.31\%} & \bf 22.95 &\emph{-13.41\%}& \bf 94.42 &\emph{+4.66\%} & \bf 93.13&\emph{+31.14\%}\\

\bottomrule   
    \end{tabular}
\end{footnotesize}
\end{center}

\end{table*}

\subsection{Settings} 

We study OOD performance using classic models and datasets for image recognition. In particular for the $f(x)$ classification function, we resort to WideResNet~\cite{ZagoruykoK2016}, which is known to provide state-of-the-art results on CIFAR and significant improvements on ImageNet, with 16-layer-deep wide residual network outperforming in accuracy  even  thousand-layer-deep networks. 
To thoroughly assess OOD performance, we therefore train WideResNet on CIFAR10~\cite{Krizhevsky09} as in-distribution, and use multiple datasets as out-of-distribution -- namely, Textures~\cite{Cimpoi2013}, SVHN~\cite{Netzer2011}, LSUN-Crop~\cite{Yu2015}, LSUN-Resize~\cite{Yu2015}, iSUN~\cite{Xu2015} that are also used in \cite{liu2021energybased}.

To carry on a fair algorithmic comparison, we do not allow fine-tuning based on OOD datasets. As such, we compare with Energy score without the fine tuning process --as otherwise the OOD data would be known at training time,  questioning whether it would be more adviseable to include such data for training-- and use the default setting proposed in  \cite{Hendrycks2017}. For our implementation, we have selected the temperature which has the smallest standard deviation on in-distribution's validation data from $T\in[1,5]$ (specifically $T=5$), and set $\epsilon=10^{-16}$.

\subsection{Metrics} 

As rightly observed in \cite{Dan2019},  OOD methods should be evaluated  on their ability to detect OOD points i.e., to focus on this capability, we consider OOD points as positive. Thus, a True Positive for the $g(x)$ function equals to  a correctly detected out-of-distribution sample (i.e., a correct rejection of a wrong $f(x)=\ell$ classification), whereas a False Positive equals to a wrongly rejected in-distribution sample (i.e., a wrong rejection of a correct $f(x)=\ell$ classification).

We then evaluate OOD detection capabilities using the  standard metrics of information retrieval, as follows.
\textbf{AUROC}  is the Area Under the Receiver Operating Characteristic (ROC) curve of False Positive Rate (FPR) and True Positive Rate (TPR) and \textbf{AUPR} is the area under the precision-recall curve. AUROC and AUPR evaluate the overall method capabilities over all operational settings. 
Additionally, we investigate the system ability to achieve high recall for a low false alarm rate by considering \textbf{TPR10}, i.e., the TPR when FPR=10$\%$.  This metric has a high practical relevance, since $g(x)$ OOD detection capabilities should not reject correct DL prediction, which would (doubly) waste computational power of $f(x)$ (and $g(x)$).
Finally, we consider the false alarm rate at high recall rate   \textbf{FPR95}, i.e., the FPR at TPR=95$\%$. Again, in practical settings, this metrics well reflect the scenario where one may wants $g(x)$ to accurately collect as much OOD samples  as possible, in order to update the $f(x)$ model.

\subsection{Results} 
\label{sec:results}
Experimental results are reported in Tab~\ref{tab:accuracy}. In general, Thinkback outperforms the SoftMax baseline (Energy score state of the art) for TPR10, FRP95, AUROC and AUPR. The average  improvement over SoftMax (Energy score) is 2.74\% (1.85\%)  for AUROC  and 25.97\% (16.67\%) for AUPR. By looking closer, at low FPR rate, Thinkback allows to detect 12.54\% (3.86\%) more OOD data than the SoftMax (Energy score). In order to detect 95\% of OOD, Thinkback  raises 8.16\% (14.03\%) less false alarm than the SoftMax baseline (Energy score). The gains are consistent across all datasets with the exception of LSUN Crop, for which Energy score provide better results.

As a side note, we report on the time complexity of the $g(x)$ methods: our  non-optimized  implementation on Pytorch v1.8, Thinkback takes 1.5 ms/sample on average (i.e., equivalent to 667\,samples/sec), which is slightly slower but still comparable to the SoftMax and Energy score (both take 1.2\,ms/sample or 833 samples/sec). 

%% file: 07_discussion.tex
We introduce Thinkback, a gradient-based method for  OOD samples  detection, fit to work with existing models (as it does not require to alter the network architecture), and in general settings (as it does not require fine-tuning on OOD data). Experimental results show Thinkback to be fast and accurate, providing superior results to the state of the art.